\PassOptionsToPackage{numbers,compress}{natbib} % MUST come first

\documentclass{article}

\usepackage[preprint]{neurips_2025}

\usepackage[utf8]{inputenc}
\usepackage[T1]{fontenc}

\usepackage{microtype}
\usepackage{xspace}
\usepackage{graphicx}
\usepackage{amsmath,amssymb,amsfonts}
\usepackage{booktabs}
\usepackage{longtable}
\usepackage{array}
\usepackage{multirow}
\usepackage{makecell}
\usepackage{multicol}
\usepackage{wrapfig}
\usepackage{subcaption}
\usepackage{enumitem}
\usepackage{titletoc}
\usepackage{pifont}
\usepackage{pdflscape}
\usepackage{listings}
\usepackage{tcolorbox}

\PassOptionsToPackage{table,x11names}{xcolor}
\usepackage{xcolor}

\usepackage{hyperref}
\hypersetup{
    colorlinks=true,
    urlcolor=blue,
    linkcolor=red,
    citecolor=blue
}

\setlist[itemize]{leftmargin=*}

\newcommand{\keywords}[1]{%
  \vspace{1em}
  \noindent\textbf{Keywords:} #1
}

\title{\bf SHIELD: Semantic Heterogeneity Integrated Embedding for Latent Discovery in Clinical Trial Safety Signals}

\author{
\textbf{Francois Vandenhende$^{1}$ \quad Anna Georgiou$^{1}$ \quad Theodoros Psaras$^{1}$ \quad Ellie Karekla$^{1}$}\\[6pt]
$^1$ClinBAY Limited, Limassol, Cyprus\\
Correspondence: \texttt{francois@clinbay.com}\\
\url{https://app.clinbay.com/safeterm}
}

\begin{document}

\maketitle

%%==================================%%
%% Abstract
%%==================================%%
\begin{abstract}
	We present SHIELD, a novel methodology for automated and integrated safety signal detection in clinical trials. SHIELD  combines disproportionality analysis with semantic clustering of adverse event (AE) terms applied to MedDRA term embeddings. For each AE, the pipeline computes an information-theoretic disproportionality measure (Information Component) with effect size derived via empirical Bayesian shrinkage. A utility matrix is constructed by weighting semantic term–term similarities by signal magnitude, followed by spectral embedding and clustering to identify groups of related AEs. 
	
	Resulting clusters are annotated with syndrome-level summary labels using large language models, yielding a coherent, data-driven representation of treatment-associated safety profiles in the form of a network graph and hierarchical tree. 
	
	We implement the SHIELD framework in the context of a single-arm incidence summary, to compare two treatment arms or for the detection of any treatment effect in a multi-arm trial.
	We illustrate its ability to recover known safety signals and generate interpretable, cluster-based summaries in a real clinical trial example. This work bridges statistical signal detection with modern natural language processing to enhance safety assessment and causal interpretation in clinical trials.
	\end{abstract}


\section{Introduction}

The evaluation of treatment-emergent adverse events (AEs) is a central component of safety assessment in clinical trials and plays a critical role in regulatory and development decision-making. Conventional safety analyses rely primarily on tabular summaries of MedDRA-coded events, structured according to the MedDRA hierarchy, and are often supplemented by aggregated summaries across pre-specified groups of terms. MedDRA provides a rich and hierarchical vocabulary for AE coding; however, it does not explicitly encode clinical knowledge beyond its predefined parent--child taxonomy \cite{ich1999_meddra_structure, mozzicato2009_meddra_overview}. Common aggregation strategies include the use of Standardised MedDRA Queries (SMQs) \cite{mozzicato2007_smq_signal, chang2017_smq_usage}, FDA Office of Custom Medical Queries \cite{fdaOCMQ_article2025}, and grouping at the System Organ Class (SOC) or hybrid levels \cite{dupuch2012_hybrid_grouping}. These tabular summaries are frequently accompanied by graphical displays such as volcano plots or forest plots \cite{atypon2013_ae_volcano, phillips2020_visualisations_ae}. 

Simple disproportionality metrics, including the Reporting Odds Ratio and Proportional Reporting Ratio, are commonly used to identify AEs with higher-than-expected incidence in one treatment arm \citep{Evans2001,Rothman2004}. While these approaches are well established and endorsed in regulatory guidance \cite{ichE2A, ichE9}, they largely treat adverse events as independent entities and provide limited support for uncovering latent structure within high-dimensional AE data or for identifying coherent safety patterns spanning related medical concepts \cite{nguyen2022_visual_harms}. Consequently, important semantic relationships between Preferred Terms (PTs) may be overlooked, particularly in trials with large and heterogeneous safety profiles, limiting the ability of reviewers to detect emerging syndromic safety signals.

A further limitation of conventional disproportionality analyses arises in the presence of rare adverse events, where small cell counts can lead to inflated or highly unstable observed-to-expected ratios. This phenomenon is particularly problematic in high-dimensional safety evaluations and in early-phase or moderate-sized clinical trials, where sparse data are common and spurious signals may be overemphasized. To address this instability, Bayesian shrinkage approaches have been proposed \cite{Bate1998, DuMouchel1999, Noren2010} that incorporate external or prior information to stabilize inference and downweight extreme estimates driven by low event counts. Among these, the Information Component (IC) has been widely adopted in post-marketing pharmacovigilance \cite{Bate1998}. The IC is formulated within an empirical Bayes framework, in which the observed-to-expected ratio of drug--event reporting frequencies, under an independence assumption, is shrunk toward a prior expectation, yielding a more robust and interpretable measure of disproportionality. Despite these favorable statistical properties, IC-based analyses are typically conducted at the level of individual drug--event pairs and do not leverage latent groupings of related adverse events that may reflect underlying clinical syndromes.

Several prior efforts have explored the use of semantic similarity, embedding-based grouping, and graphical methods to enhance adverse event analysis and safety signal detection. Dupuch et al.\ \cite{dupuch2012_hybrid_grouping} demonstrated that clustering MedDRA terms based on semantic distance can support the construction of clinically meaningful groupings for pharmacovigilance and query development. Building on the idea of embedding MedDRA concepts into vector spaces, recent work by Vandenhende et al.\ developed the SafeTerm framework, which embeds MedDRA Preferred Terms and query terms in a high-dimensional semantic space and applies cosine similarity and clustering to generate automated MedDRA queries aligned with expert-defined sets \cite{Vandenhende2025_SafeTermAMQ, Vandenhende2025_SafeTermQuery}. Extending this trajectory, a knowledge-based graphical method for safety signal detection in clinical trials was also proposed, which augments MedDRA PTs with a learned semantic knowledge layer and applies shrinkage incidence ratios and precision-weighted cluster aggregation for AE signal detection, demonstrating recovery of expected safety signals across multiple legacy trial datasets \cite{Vandenhende2025_TrialGraph}. Related consensus clustering methods, such as vigiGroup, have shown that combining empirical Bayes shrinkage with clustering can improve the stability and coherence of safety signal groupings in large spontaneous reporting databases \cite{noren2013_vigigroup}. These approaches illustrate the utility of embeddings, semantic clustering, and shrinkage for term aggregation and early signal identification. However, most existing methods have been developed in post-marketing or semi-automated query generation contexts, or require manual interpretation, and are not designed as an integrated, end-to-end pipeline for controlled clinical trial data where incidence counts, multi-arm comparisons, and syndrome interpretation must be considered jointly.

SHIELD (Semantic Heterogeneity Integrated Embedding for Latent Discovery) addresses these limitations by integrating disproportionality analysis with semantic clustering of MedDRA terms. First, we adapt an IC-like statistic to the multi-arm clinical trial setting, computing for each PT an information-theoretic measure of differential incidence across treatment arms. We then construct a graph over PTs using cosine similarity of their learned embeddings, with edges weighted by the magnitude of the IC effect. Spectral clustering of this graph, based on a normalized Laplacian followed by Ward hierarchical linkage, groups terms into clusters representing coherent treatment-emergent syndromes. Finally, each cluster is annotated using a large language model applied to its constituent terms, enabling generation of interpretable, narrative safety summaries. This integrated pipeline supports the detection and interpretation of treatment-emergent syndromic safety signals in clinical trials.

In this paper, we present the SHIELD methodology in detail. We first describe the adaptation of the Information Component for multi-arm clinical trial data, including shrinkage to stabilize rare-event estimates. Next, we outline the construction of PT embeddings and semantic similarity graphs, followed by spectral clustering using a normalized Laplacian and hierarchical linkage. We then detail the procedure for cluster annotation using large language models. To illustrate the value of SHIELD, we analyse a clinical trial example, illustrating its ability to recover expected signals, and reveal latent syndrome structure at the treatment level and in comparison to a control arm. 

\section{Methods}

\subsection{Incidence Table Construction}
The input to SHIELD is a standard incidence table of subjects with adverse events (AEs) collected from a randomized clinical trial.  
Each row corresponds to a MedDRA Preferred Term (PT), with counts of subjects experiencing that AE in each treatment arm, along with the total number of subjects at risk.  
Such tables can be constructed for all treatment-emergent AEs, or a subset (e.g., serious AEs, or events in specific sub-populations).  
Terms with zero events in all selected arms are excluded, as they provide no information for signal detection.  
Let $m$ denote the number of Preferred Terms (PTs) and $k$ the number of treatment arms. For arm $j$, let $N_j$ be the number of subjects at risk, and $c_{i,j}$ the number of subjects experiencing PT $i$. The proportion of subjects with event $i$ in group $j$ is:
\[
p_{i,j} = c_{i,j}/N_j.
\]

\subsection{Semantic Embeddings and Similarity Matrix}
To capture semantic relationships between PTs, we use precomputed normalized embeddings for MedDRA PTs, as determined in the Safeterm model \cite{Vandenhende2025_SafeTermAMQ}.  
Each PT is represented as a high-dimensional vector, where proximity in the embedding space reflects semantic similarity.  
We compute the global cosine similarity matrix among all PT embeddings as
\[
S_{\mathrm{global}}(i,j) = \frac{\mathbf{v}_i \cdot \mathbf{v}_j}{\|\mathbf{v}_i\|\,\|\mathbf{v}_j\|},
\]
where $\mathbf{v}_i$ and $\mathbf{v}_j$ are the embedding vectors for PTs $i$ and $j$, respectively.  
From this, we extract the submatrix $S'$ corresponding to the PTs observed in the trial, which will be used for semantic weighting in subsequent analyses.
\subsection{Information Component Calculation}

Let $c_{i,j}$ denote the observed number of events for preferred term (PT) $i$ in study arm $j$ ($j=1,\dots,k$), and let $N_j$ denote the total number of subjects in arm $j$. Define
\[
T_i = \sum_{j=1}^{k} c_{i,j},
\qquad
N_{\mathrm{tot}} = \sum_{j=1}^{k} N_j.
\]
All computations are performed in double precision with a small numerical constant $\varepsilon = 10^{-12}$ added where required for numerical stability.

\paragraph{Expected counts under the null.}

Under the null hypothesis of no difference across arms, the global event probability for PT $i$ is estimated as
\[
\hat p_{i,\cdot} = \frac{T_i}{N_{\mathrm{tot}}}.
\]
The expected counts are then
\[
E_{i,j} = \hat p_{i,\cdot} \, N_j.
\]
In matrix form, this corresponds to the outer product between the vector of global event probabilities and the vector of arm sample sizes.

\paragraph{Raw Information Component (IC).}

The pointwise mutual information (PMI) contribution for PT $i$ and arm $j$ is defined as
\[
\mathrm{PMI}_{i,j}
=
\log_2
\frac{c_{i,j} + \varepsilon}{E_{i,j} + \varepsilon}.
\]
Let
\[
\hat p_{j|i}
=
\frac{c_{i,j}}{T_i + \varepsilon}
\]
denote the empirical conditional probability of arm $j$ given event $i$.  
The raw Information Component is then computed as
\[
\mathrm{IC}_i
=
\sum_{j=1}^{k}
\hat p_{j|i}\,
\mathrm{PMI}_{i,j}.
\]
This expression is equivalent to the Kullback--Leibler divergence (\cite{KullbackLeibler1951}) between the empirical arm distribution given the event and the expected arm distribution under the null.

\paragraph{Frequentist likelihood-ratio test (G-test).}

For each PT, the likelihood-ratio statistic is computed as
\[
G_i
=
2\, T_i \, \mathrm{IC}_i \ln 2.
\]
Under the null hypothesis,
\[
G_i \sim \chi^2_{k-1}.
\]
Then, the upper-tail $p$-value is computed from a chi-square distribution with $k-1$ degrees of freedom, as:
\[
p_i
=
P(\chi^2_{k-1} \ge G_i).
\]

\paragraph{Directional assignment for two-arm trials.}

When $k=2$, directionality of the disproportionality measure is determined by comparing observed and expected counts in the second arm:
\[
\mathrm{sign}_i
=
\begin{cases}
+1, & c_{i,2} > E_{i,2},\\
-1, & c_{i,2} \le E_{i,2}.
\end{cases}
\]
For multi-arm comparisons ($k>2$), the IC is treated as a global, unsigned divergence measure.

\paragraph{Hierarchical Bayesian shrinkage via Dirichlet prior.}

To stabilize estimates, particularly for rare events, we adopt a hierarchical Dirichlet model for the arm probabilities.  
For each PT $i$, define the arm probability vector
\[
\boldsymbol{\pi}_i = (\pi_{i,1}, \dots, \pi_{i,k}),
\]
where $\pi_{i,j}$ represents the probability that an event of type $i$ occurs in arm $j$.

\subparagraph{Hyperprior estimation.}

Across PTs, empirical proportions
\[
\hat p_{i,j}
=
\frac{c_{i,j}}{T_i + \varepsilon}
\]
are used to estimate the hyperprior parameters via method-of-moments.  
Let $\hat\mu_j$ and $\widehat{\mathrm{Var}}_j$ denote the sample mean and variance of $\hat p_{i,j}$ across PTs. The total concentration parameter $\alpha_0$ is estimated from
\[
\alpha_0
\approx
\frac{\hat\mu_j (1 - \hat\mu_j)}{\widehat{\mathrm{Var}}_j} - 1,
\]
and the global Dirichlet hyperparameters are defined as
\[
\alpha_j = \alpha_0 \hat\mu_j.
\]
A robust median across candidate $\alpha_0$ values is used, with lower bounds enforced for numerical stability.

\subparagraph{Posterior sampling.}

Given counts $c_{i,j}$ and hyperparameters $\alpha_j$, the posterior distribution is
\[
\boldsymbol{\pi}_i \mid \mathbf{c}_i
\sim
\mathrm{Dirichlet}(c_{i,1} + \alpha_1, \dots, c_{i,k} + \alpha_k).
\]
Posterior samples are generated via Gamma sampling:
\[
\tilde \pi_{i,j}^{(s)}
=
\frac{Y_{i,j}^{(s)}}{\sum_{l=1}^{k} Y_{i,l}^{(s)}},
\qquad
Y_{i,j}^{(s)} \sim \mathrm{Gamma}(c_{i,j} + \alpha_j, 1).
\]

For each posterior draw, the IC is computed as
\[
\mathrm{IC}_i^{(s)}
=
\sum_{j=1}^{k}
\tilde \pi_{i,j}^{(s)}
\log_2
\frac{\tilde \pi_{i,j}^{(s)}}{\mu_{i,j}},
\]
where
\[
\mu_{i,j}
=
\frac{E_{i,j}}{\sum_{l=1}^{k} E_{i,l}}
\]
is the normalized expected arm distribution.

Posterior relative risk (RR) samples are also computed:
\[
\mathrm{RR}_{i,j}^{(s)}
=
\frac{\tilde \pi_{i,j}^{(s)}}{\mu_{i,j}}.
\]

\paragraph{Posterior summaries.}

For both IC and RR, posterior summaries are computed from Monte Carlo samples:

\begin{itemize}
\item posterior mean,
\item posterior median,
\item equal-tailed credible interval at confidence level $\gamma$,
\[
\left[
Q_{(1-\gamma)/2}, \,
Q_{1-(1-\gamma)/2}
\right].
\]
\end{itemize}

These summaries provide shrinkage-stabilized estimates and uncertainty quantification for each preferred term.

\paragraph{Multi-arm ($k>2$) and two-sided interpretation.}

For $k>2$, the IC represents a global divergence across arms with
\[
G_i \sim \chi^2_{k-1}.
\]
Post-hoc inspection of posterior RR samples is used to identify which arms contribute most to the deviation.  
For $k=2$, directional interpretation is based on the sign rule described above, while posterior credible intervals provide shrinkage-adjusted evidence of imbalance.
\subsection{Semantic Utility Matrix Construction}
We now build a contextualized similarity matrix $U$ that integrates both the statistical signal from the clinical trial and the semantic relationships among PTs. We use this matrix to create a network graph where nodes represent signal intensity $U_{i,i}$ at PTs and edges connect related PTS using $U_{i,j}$.

Let $S'$ denote the cosine-similarity matrix among PT embeddings. 
Weak similarities are removed by thresholding at a minimum similarity 
parameter $\tau = \texttt{sim\_min}$, yielding
\[
S_{ij} =
\begin{cases}
S'_{ij}, & \text{if } S'_{ij} \ge \tau, \\
0, & \text{otherwise}.
\end{cases}
\]
Thus, $S$ is the filtered semantic similarity matrix used in subsequent steps.

Let $n_{\mathrm{PT}}$ denote the number of preferred terms (PTs) and 
$k$ the number of treatment arms.

\paragraph{Multi-arm setting ($k>1$).}
For trials with more than one treatment arm, SHIELD computes global 
Information Component (IC) statistics using the full contingency table 
of PT counts. Let
\[
\mathbf{z} = (z_i)_{i=1}^{n_{\mathrm{PT}}}
\]
be the vector of posterior lower-bound IC estimates,
\[
z_i = \mathrm{IC}_{i}^{\mathrm{lower}},
\]
obtained from the Bayesian shrinkage procedure described in Section~2.3.
These values correspond to the lower bound of the posterior IC distribution 
and serve as conservative signal strength estimates.

Define the diagonal weighting matrix
\[
Z = \mathrm{diag}(\mathbf{z}).
\]
The semantic utility matrix is then constructed as
\[
U = Z S Z,
\qquad
U_{ij} = z_i\, z_j\, S_{ij}.
\]

\paragraph{Single-arm setting ($k=1$).}
When only one treatment arm is available, no between-arm contrast can be 
computed. In this case, SHIELD defines the signal weight directly as the 
observed incidence proportion:
\[
z_i = \frac{c_{i,1}}{N_1},
\]
where $c_{i,1}$ is the number of subjects experiencing PT $i$ and $N_1$ 
is the total number of subjects in the trial.

Let
\[
Z = \mathrm{diag}(\mathbf{z}), 
\quad \mathbf{z} = (z_i)_{i=1}^{n_{\mathrm{PT}}}.
\]
The utility matrix is again defined as
\[
U = Z S Z.
\]

\paragraph{Interpretation.}
In both settings, the construction
\[
U_{ij} = z_i\, z_j\, S_{ij}
\]
ensures that PT pairs receive high utility only when they are both 
(i) strongly weighted by statistical signal (IC lower bound or incidence 
proportion) and (ii) semantically similar. The similarity thresholding 
step prevents weakly related PTs from contributing to downstream clustering. 
This formulation provides a unified and implementation-consistent framework 
for syndrome discovery across both multi-arm and single-arm safety datasets.

\subsection{Spectral Clustering and Hierarchical Linkage}

The goal of spectral decomposition is to identify coherent, high energy subsets within the utility matrix $U$, and to partition PTs into coherent, treatment-emergent syndromes. The steps are as follows:

\begin{enumerate}
    \item Construct the normalized graph Laplacian:
    \[
    L = I - D^{-1/2} U D^{-1/2}, \quad D_{ii} = \sum_j U_{ij}.
    \]
    \item Compute eigenvalues and eigenvectors of $L$. The number of near-zero eigenvalues (the eigengap) provides an initial estimate of the number of clusters.
    \item Select the eigenvectors corresponding to the smallest eigenvalues and normalize each row to unit length. This maps PTs into a low-dimensional space where Euclidean distances capture both statistical and semantic similarity.
    \item Apply hierarchical agglomerative clustering (Ward linkage) to the spectral embedding. Ward linkage (\cite{Ward01031963}) minimizes intra-cluster variance, producing compact clusters of semantically related and statistically enriched PTs.
    \item Determine the dendrogram cut point adaptively by identifying the largest gap in successive linkage distances, thereby estimating the number of clusters without arbitrary thresholds.
\end{enumerate}

\paragraph{Rationale:} The two-stage approach—spectral embedding followed by hierarchical clustering—combines flexibility and interpretability. Spectral embedding allows the detection of non-convex and complex cluster shapes that would be missed by direct clustering on raw IC values. Hierarchical clustering provides an interpretable dendrogram, facilitating visualization and annotation. The resulting clusters represent PTs that are both semantically coherent and enriched for treatment effects, forming the basis for downstream syndrome labeling and narrative summarization.

\subsection{Cluster Labeling and Reporting}

For interpretability, each cluster is annotated using a large language model (LLM). The PTs in a cluster are supplied to the LLM, which returns a unifying medical concept (e.g., “Hepatobiliary disorder”). The process is repeated iteratively across the entire clustering hierarchy.

For reporting and interpretability purpose, we generate:
\begin{itemize}
    \item A summary table listing each PT, its cluster label, incidence summaries by treatment, the adjusted fold-change ($2^{\mathrm{IC}_{\mathrm{adj}}}$), and posterior summaries.
    \item A static visualization including dendrograms with branch colors indicating clusters and bars showing effect magnitude.
    \item An interactive Network graph of the utility matrix, with effect-size thresholding capabilities.
\end{itemize}

\section{Application to Clinical Trial Data}
We applied the SHIELD methodology to a clinical trial dataset, illustrating its usefulness in single-arm summary, two-arm comparisons and multi-arm analyses.

This phase III trial \cite{NCT05096221} is a two-treatment, two-part study in Duchenne muscular dystrophy. The incidence of patients experiencing adverse events (AEs) is summarized for each treatment and part separately in Table~\ref{tab:dmd}. Table~\ref{tab:dmd} also presents the disproportionality statistics for the $k=4$ arm comparison, including the adjusted Information Component (IC) and its confidence interval, the raw signal ratio, and the $p$-value from the G-test.

Up to eight preferred term (PT) clusters were detected by SHIELD, ranging from 2 to 16 PTs per cluster. The largest clusters correspond to \emph{Liver disease} and \emph{Infectious disease}. The automated grouping of AEs observed in the trial facilitates a more comprehensive and structured safety assessment.

\begin{figure}
	\centering
	\includegraphics[width=1.0\textwidth]{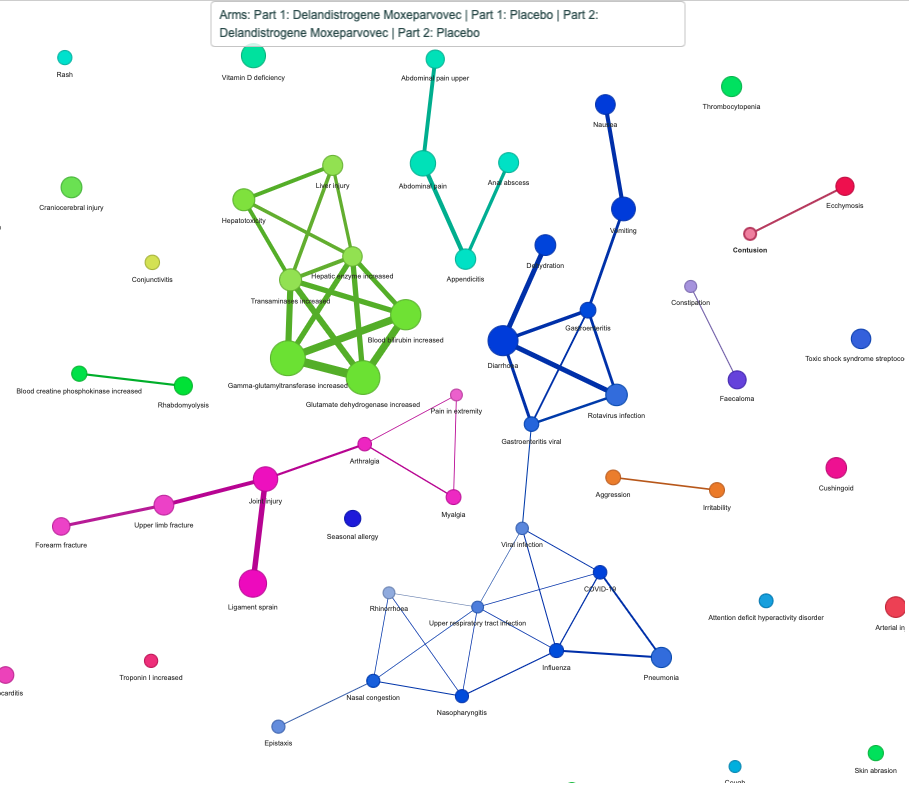}
	\caption{SHIELD network graph of AEs for the 4-arm disproportionality analysis of the DMD trial.}
	\label{fig:fig_01}
\end{figure}

Figure~\ref{fig:fig_01} presents the SHIELD network graph for the four-arm comparison. Node size is proportional to $IC_{2.5\%}$, highlighting, among other findings, a disproportionate incidence of AEs between arms within the \emph{Liver disease} cluster.

\begin{figure} 
	\centering
	\includegraphics[width=\textwidth,height=\textheight,keepaspectratio]{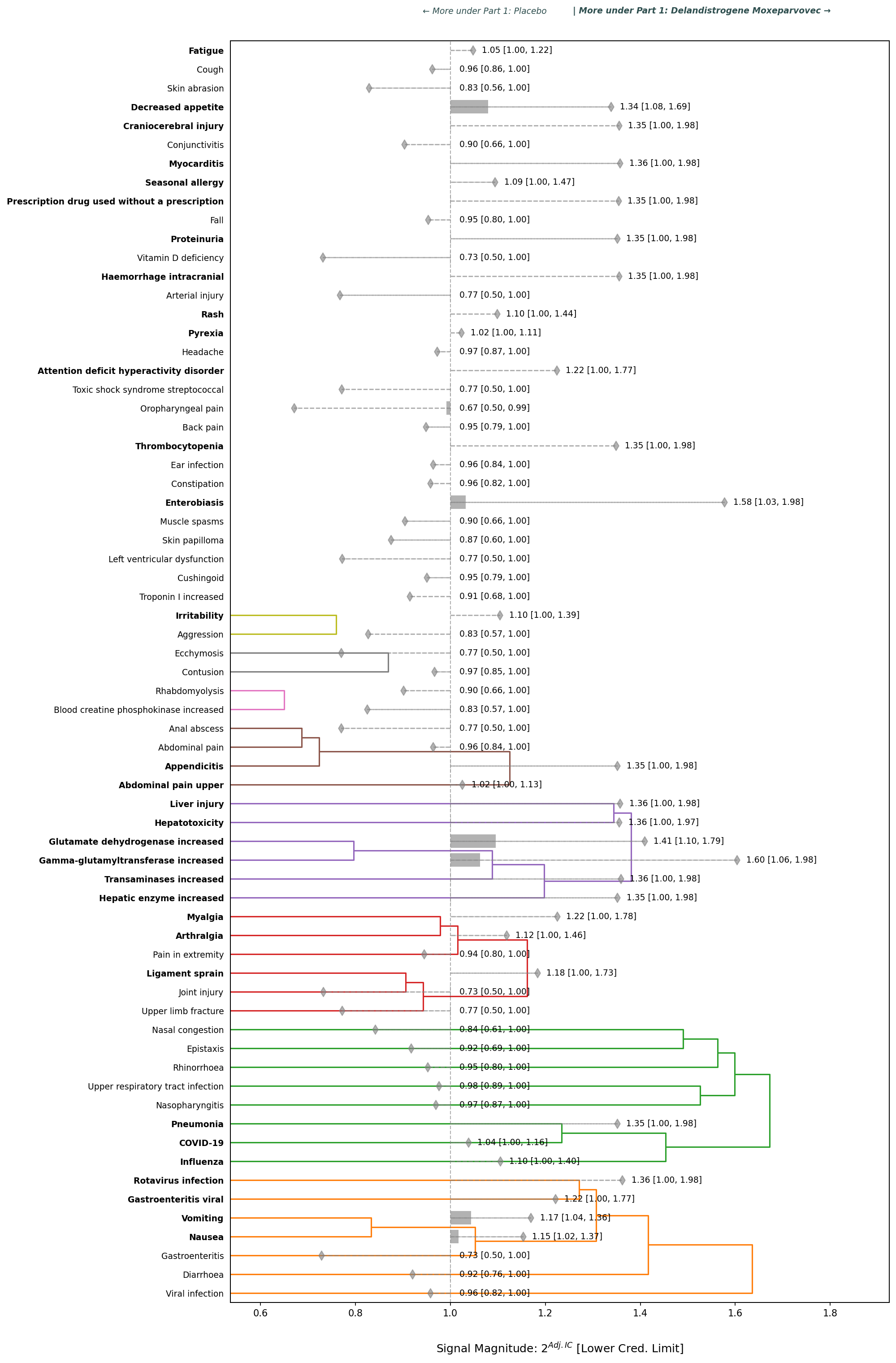}
	\caption{Dendogram plot of AE clusters for the 2-arm disproportionality analysis of the DMD trial during Part 1.}
	\label{fig:fig_02}
\end{figure}

When comparing the $k=2$ treatments in Part 1, Figure~\ref{fig:fig_02} shows the resulting dendrogram for PT grouping together with the signed disproportionality statistic and its lower $2.5\%$ bound. A significant increase in liver-related AEs is observed in the active treatment arm, with all PTs in the \emph{Liver disease} cluster exhibiting a consistent signal of disproportionality.

\begin{figure}
	\centering
	\includegraphics[width=1.0\textwidth]{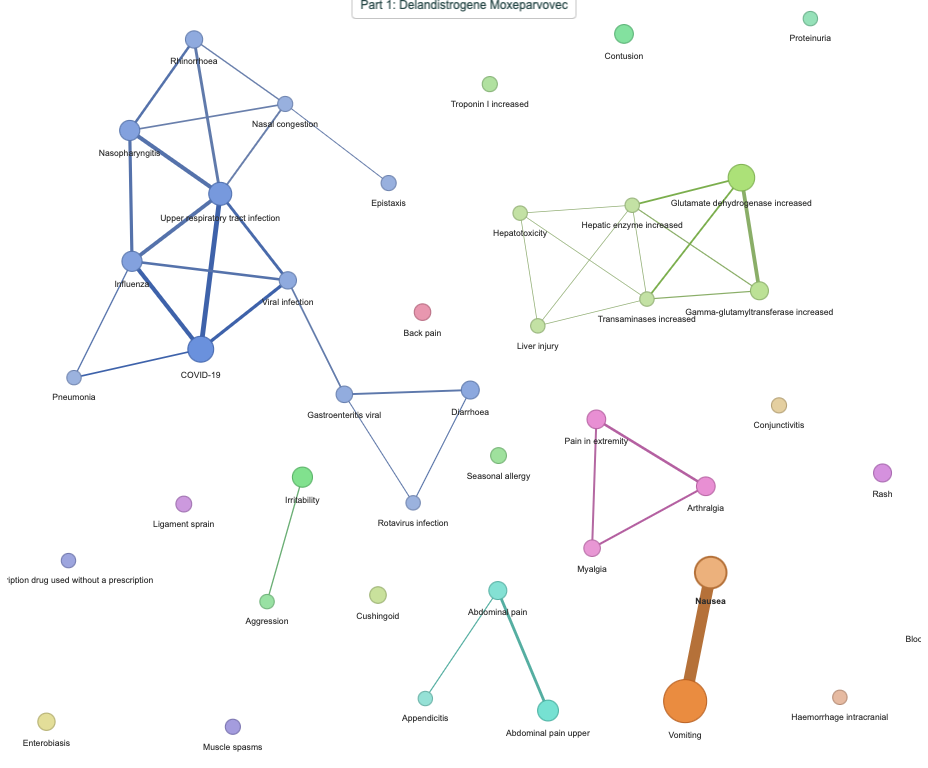}
	\caption{SHIELD network graph of AEs for the single-arm incidence analysis of the DMD trial.}
	\label{fig:fig_03}
\end{figure}

Finally, Figure~\ref{fig:fig_03} presents the SHIELD network graph for the active arm in Part 1 ($k=1$), providing a single-arm summary of observed AE incidences. The figure shows similar AE groupings as in the disproportionality analyses, but with different node and edge weights, since the signal is based on observed incidence rather than treatment contrasts. This visualization provides an intuitive overview of the safety profile in the active arm, highlighting clusters of related AEs and their relative frequencies.

	\section{Discussion}

	This study demonstrates the potential of using contextualized network graphs for enhanced safety signal detection in clinical trials. It combines semantic similarity of MedDRA terms with treatment comparison statistics to identify clusters of related AEs that may represent underlying syndromes. The method was applied to a Duchenne Muscular Dystrophy trial, where it successfully identified a cluster of liver-related AEs with increased incidence in the active treatment arm, consistent with known drug effects. 
	
	While these results are promising, further work is needed to systematically quantify performance of the network graph for safety signal detection across a broader range of clinical trials. 
	Future research could incorporate additional AE attributes, such as severity, duration, and onset time. Individual patient-level data could also be analyzed, including repeated- or co-occurences of events over time.

	Last, the method could be generalised to other clinical data types, such as laboratory parameters, ECG, vital signs, survival or other efficacy endpoints, if they are supported by proper MedDRA terminology.
	
	\section{Conclusion}

The SHIELD network graph created using contextualized similarity grouping of adverse events is helpful to streamline safety signal detection in clinical trials.
It provides an accessible, interactive platform for exploring trial AE data, supporting more informed safety review. The method combining statistical inference with semantic clustering represents a novel 
approach to enhance the interpretability and clinical relevance of safety analyses in clinical trials. The SHIELD methodology is available as an application programming interface (API) and a web interface at \url{https://app.clinbay.com/safeterm}.

\section*{Statements and Declarations}

\textbf{Conflicts of interest:} 
The authors are affiliated with ClinBAY Ltd; they declare no financial or commercial conflict beyond this.\\[4pt]

\textbf{Ethics approval:} Not applicable; retrospective analysis of existing data.\\[4pt]

\textbf{Data availability:}\\
\url{https://clinicaltrials.gov/study/NCT05096221}\\

\bibliographystyle{unsrtnat}
\bibliography{main}
\clearpage

\clearpage
\begin{landscape}
	\begin{longtable}{p{3cm}p{3.5cm}p{1.5cm}p{2.0cm}p{1.5cm}p{1.5cm}p{2.0cm}p{2.0cm}p{2.0cm}p{2.0cm}}
    \caption{Incidence of Subjects with AE by Preferred Terms grouped by Cluster with Adjusted Signal Estimates} \label{tab:dmd} \\
	\toprule
    Cluster &
    PT &
    Adjusted Signal &
    Adjusted Signal (CI) &
    Raw Signal Ratio &
    $p$-value &
    Part 1: Delandistrogene Moxeparvovec &
    Part 1: Placebo &
    Part 2: Delandistrogene Moxeparvovec &
    Part 2: Placebo \\
    \midrule
    \endfirsthead
    
    \toprule
    Cluster &
    PT &
    Adjusted Signal &
    Adjusted Signal (CI) &
    Raw Signal Ratio &
    $p$-value &
    Part 1: Delandistrogene Moxeparvovec &
    Part 1: Placebo &
    Part 2: Delandistrogene Moxeparvovec &
    Part 2: Placebo \\
    \midrule
    \endhead
	\multirow[t]{30}{*}{None} & Myocarditis & 1.49 & (1.05, 2.46) & 1.97 & 0.4388 & 1/63 & 0/62 & 0/60 & 1/63 \\
	 & Ketonuria & 2.05 & (1.20, 3.64) & 4.13 & 0.0365 & 0/63 & 0/62 & 3/60 & 0/63 \\
	 & Rash & 1.21 & (1.03, 1.52) & 1.15 & 0.1580 & 6/63 & 3/62 & 8/60 & 2/63 \\
	 & Toxic shock syndrome streptococcal & 1.67 & (1.08, 3.07) & 4.00 & 0.4280 & 0/63 & 1/62 & 0/60 & 0/63 \\
	 & Skin papilloma & 1.52 & (1.08, 2.40) & 1.57 & 0.2101 & 1/63 & 1/62 & 3/60 & 0/63 \\
	 & Fatigue & 1.25 & (1.07, 1.48) & 1.21 & 0.0177 & 9/63 & 6/62 & 10/60 & 1/63 \\
	 & Cough & 1.06 & (1.01, 1.15) & 1.03 & 0.2432 & 12/63 & 19/62 & 20/60 & 12/63 \\
	 & Left ventricular dysfunction & 1.68 & (1.08, 3.13) & 4.00 & 0.4280 & 0/63 & 1/62 & 0/60 & 0/63 \\
	 & Thrombocytopenia & 1.56 & (1.09, 2.51) & 1.71 & 0.0927 & 1/63 & 0/62 & 4/60 & 1/63 \\
	 & Seasonal allergy & 1.31 & (1.05, 1.73) & 1.37 & 0.1308 & 3/63 & 2/62 & 0/60 & 4/63 \\
	 & Muscle spasms & 1.17 & (1.01, 1.54) & 1.06 & 0.7806 & 2/63 & 4/62 & 2/60 & 2/63 \\
	 & Pyrexia & 1.12 & (1.03, 1.25) & 1.10 & 0.0178 & 20/63 & 15/62 & 14/60 & 5/63 \\
	 & Troponin I increased & 1.18 & (1.02, 1.49) & 1.13 & 0.3131 & 2/63 & 2/62 & 5/60 & 6/63 \\
	& Back pain & 1.08 & (1.01, 1.25) & 1.01 & 0.9669 & 4/63 & 4/62 & 5/60 & 5/63 \\
	 & Arterial injury & 1.69 & (1.09, 3.07) & 4.00 & 0.4280 & 0/63 & 1/62 & 0/60 & 0/63 \\
	 & Fall & 1.07 & (1.00, 1.22) & 1.01 & 0.9094 & 5/63 & 7/62 & 6/60 & 5/63 \\
	 & Proteinuria & 1.41 & (1.05, 2.06) & 1.54 & 0.1116 & 1/63 & 0/62 & 2/60 & 4/63 \\
	 & Vitamin D deficiency & 1.64 & (1.13, 2.45) & 2.08 & 0.0621 & 0/63 & 2/62 & 3/60 & 0/63 \\
	 & Prescription drug used without a prescription & 1.72 & (1.09, 3.17) & 3.94 & 0.4334 & 1/63 & 0/62 & 0/60 & 0/63 \\
	 & Oropharyngeal pain & 1.35 & (1.05, 1.89) & 1.42 & 0.1321 & 0/63 & 4/62 & 2/60 & 2/63 \\
	& Skin abrasion & 1.34 & (1.04, 2.08) & 1.26 & 0.3536 & 1/63 & 4/62 & 1/60 & 1/63 \\
	 & Cushingoid & 1.36 & (1.09, 1.68) & 1.35 & 0.0863 & 4/63 & 4/62 & 3/60 & 0/63 \\
	  & Craniocerebral injury & 1.74 & (1.10, 3.20) & 3.94 & 0.4334 & 1/63 & 0/62 & 0/60 & 0/63 \\
	 & Decreased appetite & 1.48 & (1.22, 1.77) & 1.48 & 0.0000 & 20/63 & 3/62 & 15/60 & 1/63 \\
	 & Enterobiasis & 1.44 & (1.10, 1.91) & 1.55 & 0.0215 & 5/63 & 0/62 & 1/60 & 5/63 \\
	& Ear infection & 1.15 & (1.02, 1.38) & 1.08 & 0.4360 & 6/63 & 6/62 & 4/60 & 2/63 \\
	 & Headache & 1.05 & (1.00, 1.14) & 1.01 & 0.9133 & 8/63 & 8/62 & 10/60 & 10/63 \\
	 & Haemorrhage intracranial & 1.73 & (1.10, 3.17) & 3.94 & 0.4334 & 1/63 & 0/62 & 0/60 & 0/63 \\
	 & Conjunctivitis & 1.29 & (1.03, 1.84) & 1.19 & 0.4331 & 2/63 & 4/62 & 1/60 & 1/63 \\
	 & Attention deficit hyperactivity disorder & 1.26 & (1.03, 1.72) & 1.18 & 0.4030 & 4/63 & 1/62 & 1/60 & 3/63 \\
	 \multirow[t]{2}{*}{Trauma} & Ecchymosis & 1.53 & (1.07, 2.46) & 1.71 & 0.0915 & 0/63 & 1/62 & 4/60 & 1/63 \\
	 & Contusion & 1.08 & (1.01, 1.26) & 1.03 & 0.6613 & 7/63 & 9/62 & 5/60 & 5/63 \\
	 \multirow[t]{2}{*}{Rhabdomyolysis} & Rhabdomyolysis & 1.44 & (1.07, 2.09) & 1.53 & 0.1150 & 2/63 & 4/62 & 0/60 & 1/63 \\
	 & Blood creatine phosphokinase increased & 1.34 & (1.04, 2.08) & 1.26 & 0.3536 & 1/63 & 4/62 & 1/60 & 1/63 \\
	 \multirow[t]{2}{*}{Bowel Obstruction} & Faecaloma & 1.57 & (1.07, 2.81) & 3.94 & 0.4334 & 0/63 & 0/62 & 0/60 & 1/63 \\
	 & Constipation & 1.08 & (1.01, 1.25) & 1.01 & 0.9171 & 5/63 & 5/62 & 6/60 & 4/63 \\
	 \multirow[t]{2}{*}{Impulse Control Disorder} & Irritability & 1.23 & (1.04, 1.61) & 1.17 & 0.1270 & 9/63 & 4/62 & 3/60 & 2/63 \\
	 & Aggression & 1.27 & (1.03, 1.69) & 1.21 & 0.1706 & 1/63 & 4/62 & 6/60 & 2/63 \\
	 \multirow[t]{4}{*}{Gastrointestinal Disorders} & Abdominal pain upper & 1.24 & (1.07, 1.42) & 1.21 & 0.0082 & 10/63 & 9/62 & 11/60 & 1/63 \\
	 & Abdominal pain & 1.35 & (1.14, 1.56) & 1.35 & 0.0053 & 6/63 & 7/62 & 8/60 & 0/63 \\
	 & Appendicitis & 1.72 & (1.09, 3.24) & 3.94 & 0.4334 & 1/63 & 0/62 & 0/60 & 0/63 \\
	 & Anal abscess & 1.69 & (1.09, 3.12) & 4.00 & 0.4280 & 0/63 & 1/62 & 0/60 & 0/63 \\
	 \multirow[t]{7}{*}{Musculoskeletal trauma} & Pain in extremity & 1.06 & (1.00, 1.18) & 1.02 & 0.6244 & 7/63 & 12/62 & 10/60 & 8/63 \\
	 & Myalgia & 1.30 & (1.04, 1.79) & 1.22 & 0.2661 & 4/63 & 1/62 & 4/60 & 1/63 \\
	 & Ligament sprain & 1.57 & (1.16, 2.19) & 1.65 & 0.0180 & 3/63 & 1/62 & 6/60 & 0/63 \\
	 & Joint injury & 1.65 & (1.13, 2.51) & 2.08 & 0.0621 & 0/63 & 2/62 & 3/60 & 0/63 \\
	 & Upper limb fracture & 1.68 & (1.09, 3.11) & 4.00 & 0.4280 & 0/63 & 1/62 & 0/60 & 0/63 \\
	 & Arthralgia & 1.17 & (1.02, 1.43) & 1.11 & 0.2351 & 7/63 & 3/62 & 8/60 & 3/63 \\
	 & Forearm fracture & 1.56 & (1.06, 2.78) & 3.94 & 0.4334 & 0/63 & 0/62 & 0/60 & 1/63 \\
	 \multirow[t]{7}{*}{Liver disease} & Hepatic enzyme increased & 1.59 & (1.08, 2.59) & 2.02 & 0.4225 & 1/63 & 0/62 & 1/60 & 0/63 \\
	 & Glutamate dehydrogenase increased & 1.51 & (1.23, 1.85) & 1.51 & 0.0000 & 18/63 & 2/62 & 12/60 & 1/63 \\
	 & Hepatotoxicity & 1.66 & (1.11, 2.71) & 2.15 & 0.2037 & 1/63 & 0/62 & 2/60 & 0/63 \\
	 & Transaminases increased & 1.73 & (1.11, 3.21) & 3.94 & 0.4334 & 1/63 & 0/62 & 0/60 & 0/63 \\
	 & Gamma-glutamyltransferase increased & 1.60 & (1.24, 2.11) & 1.68 & 0.0001 & 6/63 & 0/62 & 13/60 & 2/63 \\
	 & Blood bilirubin increased & 2.04 & (1.19, 3.65) & 4.13 & 0.0365 & 0/63 & 0/62 & 3/60 & 0/63 \\
	 & Liver injury & 1.61 & (1.09, 2.63) & 2.02 & 0.4225 & 1/63 & 0/62 & 1/60 & 0/63 \\
	 \multirow[t]{16}{*}{Infectious Disease} & Nasal congestion & 1.15 & (1.02, 1.42) & 1.10 & 0.3082 & 2/63 & 7/62 & 6/60 & 4/63 \\
	  & Epistaxis & 1.21 & (1.02, 1.59) & 1.08 & 0.6915 & 2/63 & 3/62 & 3/60 & 1/63 \\
	   & Rhinorrhoea & 1.06 & (1.00, 1.19) & 1.01 & 0.9273 & 5/63 & 7/62 & 6/60 & 7/63 \\
	  & Pneumonia & 1.72 & (1.09, 3.21) & 3.94 & 0.4334 & 1/63 & 0/62 & 0/60 & 0/63 \\
	  & COVID-19 & 1.13 & (1.02, 1.32) & 1.10 & 0.0566 & 17/63 & 11/62 & 6/60 & 6/63 \\
	  & Influenza & 1.20 & (1.03, 1.49) & 1.14 & 0.1642 & 9/63 & 4/62 & 5/60 & 2/63 \\
	  & Nasopharyngitis & 1.14 & (1.02, 1.34) & 1.09 & 0.1454 & 9/63 & 12/62 & 5/60 & 4/63 \\
	  & Upper respiratory tract infection & 1.05 & (1.00, 1.14) & 1.02 & 0.5784 & 13/63 & 17/62 & 13/60 & 10/63 \\
	  & Viral infection & 1.12 & (1.01, 1.32) & 1.05 & 0.6433 & 5/63 & 5/62 & 2/60 & 5/63 \\
	  & Rotavirus infection & 1.73 & (1.10, 3.19) & 3.94 & 0.4334 & 1/63 & 0/62 & 0/60 & 0/63 \\
	  & Diarrhoea & 1.40 & (1.19, 1.65) & 1.41 & 0.0003 & 6/63 & 13/62 & 8/60 & 0/63 \\
	  & Dehydration & 1.67 & (1.10, 2.83) & 2.33 & 0.0801 & 0/63 & 0/62 & 3/60 & 1/63 \\
	  & Gastroenteritis viral & 1.35 & (1.03, 2.03) & 1.25 & 0.3674 & 4/63 & 1/62 & 1/60 & 1/63 \\
	  & Gastroenteritis & 1.32 & (1.04, 1.75) & 1.39 & 0.1152 & 0/63 & 2/62 & 3/60 & 4/63 \\
	  & Vomiting & 1.24 & (1.13, 1.39) & 1.23 & 0.0000 & 41/63 & 12/62 & 45/60 & 9/63 \\
	  & Nausea & 1.22 & (1.09, 1.40) & 1.20 & 0.0000 & 25/63 & 8/62 & 23/60 & 5/63 \\
	
      \bottomrule
    \end{longtable}
\end{landscape}
    \clearpage
\end{document}